\documentclass[manuscript,screen]{acmart}

\usepackage{times}
\usepackage{epsfig}
\usepackage{graphicx}
\usepackage{amsmath}
\usepackage{subfigure}
\usepackage{caption}
\usepackage{latexsym}
\usepackage{color}

\AtBeginDocument{%
  \providecommand\BibTeX{{%
    \normalfont B\kern-0.5em{\scshape i\kern-0.25em b}\kern-0.8em\TeX}}}

\copyrightyear{2021}
\setcopyright{acmcopyright}
\acmJournal{TOMM}
\acmYear{2021} \acmVolume{0} \acmNumber{0} \acmArticle{0} \acmMonth{0} \acmPrice{15.00}\acmDOI{00.0000/0000000}

\begin{document}

\title{FasterPose: A Faster Simple Baseline for Human Pose Estimation}


\author{Hanbin Dai}
\email{daihanbin.ac@gmail.com}
\affiliation{%
  \institution{JD AI Research}
  \city{Beijing}
  \country{China}
}

\author{Hailin Shi}
\email{shihailin@jd.com}
\affiliation{%
  \institution{JD AI Research}
  \city{Beijing}
  \country{China}
}

\author{Wu Liu}
\email{liuwu1@jd.com}
\affiliation{%
  \institution{JD AI Research}
  \city{Beijing}
  \country{China}
}

\author{Linfang Wang}
\email{wanglinfang@jd.com}
\affiliation{%
  \institution{JD AI Research}
  \city{Beijing}
  \country{China}
}

\author{Yinglu Liu}
\email{liuyinglu1@jd.com}
\affiliation{%
  \institution{JD AI Research}
  \city{Beijing}
  \country{China}
}

\author{Tao Mei}
\email{tmei@live.com}
\affiliation{%
  \institution{JD AI Research}
  \city{Beijing}
  \country{China}
}

\authorsaddresses{}

\renewcommand\footnotetextcopyrightpermission[1]{} 

\renewcommand{\shortauthors}{}

\begin{abstract}
The performance of human pose estimation depends on the spatial accuracy of keypoint localization.
Most existing methods pursue the spatial accuracy through learning the high-resolution (HR) representation from input images. By the experimental analysis, we find that the HR representation leads to a sharp increase of computational cost, while the accuracy improvement remains marginal compared with the low-resolution (LR) representation. 
In this paper, we propose a design paradigm for cost-effective network with LR representation for efficient pose estimation, named FasterPose.
Whereas the LR design largely shrinks the model complexity, yet how to effectively train the network with respect to the spatial accuracy is a concomitant challenge.
We study the training behavior of FasterPose, and formulate a novel regressive cross-entropy (RCE) loss function for accelerating the convergence and promoting the accuracy.
The RCE loss generalizes the ordinary cross-entropy loss from the binary supervision to a continuous range, thus the training of pose estimation network is able to benefit from the sigmoid function. 
By doing so, the output heatmap can be inferred from the LR features without loss of spatial accuracy, while the computational cost and model size has been significantly reduced.
Compared with the previously dominant network of pose estimation, our method reduces 58\% of the FLOPs and simultaneously gains 1.3\% improvement of accuracy.
Extensive experiments show that FasterPose yields promising results on the common benchmarks, \textit{i.e.},~ COCO and MPII, consistently validating the effectiveness and efficiency for practical utilization, especially the low-latency and low-energy-budget applications in the non-GPU scenarios.
\end{abstract}



\maketitle

\begin{figure}[t]
\centering
\includegraphics[scale=0.085]{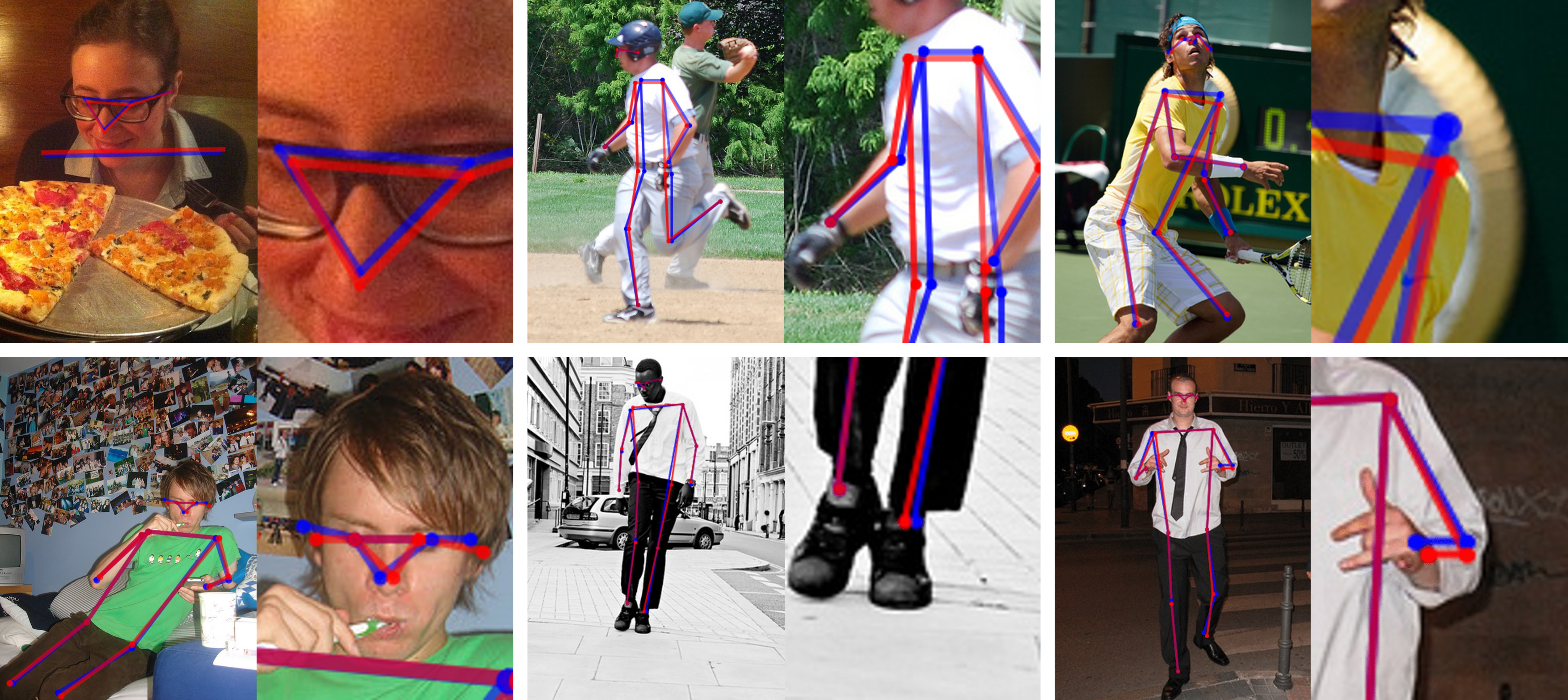}
\caption{
Qualitative evaluation of FasterPose on COCO. The red one is predicted by FasterPose, and the blue reference is by a state-of-the-art fast method, \textit{i.e.},~ SimpleBaseline.
The left side is original image, and the right side is zoomed image. FasterPose deals well with various adverse factors, such as twisted limb, abnormal pose, and view point change. It is worth noting that these results are yielded by FasterPose with 133 fps on a single NIVDIA 2080ti GPU.
}
\label{coco}
\end{figure}

\section{Introduction}
\label{sec:intro}

Human pose estimation in image and video has been a long-standing yet challenging task in computer vision.
The goal is to localize human anatomical keypoints (wrists, knees, elbows, \textit{etc}.) or parts in the input images. 
This paper is interested in single-person pose estimation, which has been applied in many practical scenarios such as
action recognition~\cite{wang2013approach,cheron2015p}, pose tracking~\cite{cho2013adaptive}, human-computer interaction~\cite{shotton2011real}, \textit{etc}. 

The recent literature shows that deep convolutional neural network (CNN) greatly improves the state-of-the-art performance in human pose estimation.
Many CNN-based methods are developed from Part-Detector~\cite{tompson2014joint} that regresses the heatmaps corresponding to the keypoints.
In the regime of heatmap regression, one can find the high-resolution (HR)  feature map provides great support to accurate keypoint localization. 
Thus, many state-of-the-art methods pursue the spatial accuracy through learning the HR representation. They~\cite{newell2016stacked,xiao2018simple,insafutdinov2016deepercut,yang2017learning,cai2020learning} dispose the network with the high-to-low resolution subnetworks, and subsequently raise the resolution in the output stage.
For instance, Hourglass~~\cite{newell2016stacked} and RSN~~\cite{cai2020learning} recovers the HR features through a symmetric low-to-high resolution process.
SimpleBaseline~\cite{xiao2018simple} adopts deconvolution layers for generating HR representations.
The dilated convolution plays a similar role in~\cite{insafutdinov2016deepercut,yang2017learning}.
In addition, HRNet~\cite{sun2019deep} maintains the HR representations throughout the whole network.

These HR feature based methods have achieved great progress for human pose estimation in recent years.
However, as shown in the experimental analysis (Section~\ref{sec:exp:ablation}), we find that the HR representation leads to a sharp increase of computational cost, while the accuracy improvement remains marginal.
For example, when the feature resolution increases from 8$\times$6 to 64$\times$48, the computational cost increases from 3.8 to 9.0 GFLOPs, while the AP merely increases by less than 0.1\% on COCO (Table~\ref{tbl:resolution}).

Therefore, we put forward a question: is there a cost-effective solution that boosts the accuracy without any additional burden of computational complexity or even with lighter weights?
To deal with this challenge, we start from our design paradigm for efficient pose estimation, named \textbf{FasterPose}.
The basic idea is to establish the thorough utilization of the low-resolution (LR) features.
FasterPose consists of an off-the-shelf backbone with high-to-low resolution, and a Low-to-High Regressor (LHR) module in succession. The design paradigm is particularly simple in structure and complexity. The backbone from any image classification routine can be directly used in FasterPose without any modification.
LHR adopts a group of 1$\times$1 convolution to each pixel of the LR feature maps that are output from the backbone to obtain multiple output values (\textit{e.g.}~ $64$), and then these multiple values are reshaped to form a region with a certain area (\textit{e.g.}~ 8$\times$8). After all the pixel points of LR feature maps are magnified many times by this operation, the LR feature maps is also magnified many times to obtain the HR heatmaps.
This LHR routine conducts more efficient inference than the deconvolution used in SimpleBaseline~\cite{xiao2018simple} which is currently the representative method for efficient post estimation. Given the same backbone, we find that the model size of FasterPose (ResNet-50) is 75\% of SimpleBaseline, and the computational complexity (FLOPs) is only 42\%.
This finding confirms that the exploitation of the LR feature is greatly useful towards the efficient human pose estimation, which nevertheless has been overlooked before. 

Based on the efficient architecture of FasterPose, 
the following objective is to effectively train the network and yield the accurate prediction for pose estimation.
Through the comparison analysis, we find the core of the problem is that the network with LR feature suffers from the slow converges, and consequently the accuracy is degraded (Fig.~\ref{fig:training_curve}).
To address the convergence problem, we inspect the distribution of the predicted values and the ground truth values on heatmap. For each person to be estimated, the true keypoints (\textit{i.e.},~ the positive samples with values greater than zero) are distributed sparsely and strongly correlated in the spatial dimension, while there are massive background pixels (\textit{i.e.},~ the negative samples with zero value) that fill the remaining space. 
When the feature resolution changes from high to low, the parameters of the heatmap regression are dramatically reduced, most of which are dedicated to the negative sample regression under the supervision of mean squared error (MSE) loss. This causes the slow convergence on the positive sample regression that precisely forms the essential objective of pose estimation.
Given the massive negative samples with zero value, the cross-entropy (CE) style supervision is a more suitable choice, as it releases the parameters for the positive regression (Section~\ref{sec:method:rce}). 
The ordinary cross-entropy loss, however, cannot be utilized to supervise the learning of pose estimation in a straightforward manner, since the ground truth of heatmap regression consists of real numbers that ranges from 0 to 1, which does not match the binary target of cross-entropy supervision. To deal with this issue, we reshape the cross-entropy loss to a novel formulation, named \textbf{Regressive Cross-Entropy} (RCE), which is able to supervise the regression toward real numbers between 0 and 1. Moreover, we find certain fine properties of RCE loss, such as the adaptive weighting of hard sample between positives and negatives. Benefiting from these advantages, RCE loss significantly improves the convergence speed and accuracy performance of FasterPose.
 
In summary, our contribution includes:
\begin{itemize}
\item 
It is the first attempt to analyze the impact of feature resolution in human pose estimation. The experimental results demonstrate that the LR features are also very useful and more efficient compared with the HR ones.
\item
Resorting to the advantage of LR feature, we propose a design paradigm for cost-effective network of pose estimation, \textit{i.e.},~ FasterPose. Compared with the HR-based counterparts, FasterPose largely reduce the computational complexity, including the inference time, model size, and FLOPs, and thereby facilitates the low-latency and low-energy-budget applications on the non-GPU devices. 
\item 
We formulate a novel loss function for heatmap regression, named RCE loss~\footnote{The source code are released at \url{https://github.com/hbin-ac/FasterPose}}, which is able to speed up the convergence of FasterPose and outperforms the counterparts in terms of test accuracy. These advantages are fully validated in the comprehensive experiments.
\end{itemize}

\section{Related Work}
\label{sec:related}

\subsection{Feature Resolution in Pose Estimation}
\label{sec:related:resolution}

HR features provide rich spatial information, many existing methods~\cite{chen2018cascaded,xiao2018simple,newell2016stacked,tang2018deeply,ke2018multi,sun2019deep,cai2020learning} pursue the accuracy through learning the HR representation. 
The representative architectures can be categorized in three subsets: 
(1) the backbone is followed by a low-to-high process (\textit{e.g.}~ bilinear-upsampling or deconvolutions) to output HR representations, such as Hourglass~\cite{newell2016stacked}, CPN~\cite{chen2018cascaded}, RSN~\cite{cai2020learning}, and SimpleBaseline~\cite{xiao2018simple};
(2) the backbone is combined with dilated convolutions~\cite{lifshitz2016human,insafutdinov2016deepercut,pishchulin2016deepcut};
(3) the backbone maintains the HR representation throughout the network, such as HRNet~\cite{sun2019deep}.
In fact, we find that the HR representation leads to a sharp increase of computational cost, while the accuracy improvement remains marginal.
Till now, there are few tools besides the bilinear interpolation and deconvolution, that dedicated to the effective upsampling from LR features for efficient pose estimation. Therefore, only one method SimpleBaseline~\cite{xiao2018simple} carries out pose estimation that utilizes mere the LR features.
In the field of super-resolution, ESPCN~\cite{shi2016real} propose the PixelShuffle routine, however, which cannot be directly utilized for pose estimation, since it is incompatible with the keypoint configuration.

\subsection{Supervision for Heatmap Regression}
\label{sec:related:supervision}

The MSE (mean square error) loss is the most widely used supervision in heatmap-based pose estimation~\cite{he2016deep,cao2017realtime,chen2018cascaded,papandreou2017towards,insafutdinov2016deepercut,Zhang_2020_CVPR}.
It optimizes the pixel-wise similarity between the output and  ground truth heatmap subject to the Euclidean metric. 
However, the training objective of MSE is not consistent with the evaluation metric of pose estimation.
The discrepancy mainly comes from that the a prior distributions of keypoint and background pixels have significant gap. One can refer to Section~\ref{sec:method:rce:mse} for the detailed discussion.
In such situation, the CE style supervision is more suitable for learning the regression towards 0 and 1. 
However, the ground truth values are real numbers that range from 0 to 1, which is incompatible with the binary objective of CE loss.
PoseFix~\cite{Moon2019PoseFix}, G-RMI~\cite{papandreou2017towards} and UDP~\cite{Huang_2020_CVPR} deal with this issue by the alternative of one-hot or mask. 
The experiments show that such modification on heatmap will cause bias and performance drop in Section~\ref{sec:exp:ablation}.
To deal with this obstacle, we reshape the CE loss to a novel formulation that supervises the regression towards real numbers between 0 and 1.

\section{Proposed Method}
\label{sec:method}

In this section, we first describe the design paradigm for efficient network, and its essential element, \textit{i.e.},~ low-to-high regressor; then, we discuss the limitation of MSE and plain CE loss in condition of LR feature, and present the formulation and application of the RCE supervision.

\begin{figure*}[t]
    \centering
    \includegraphics[width=0.9\linewidth]{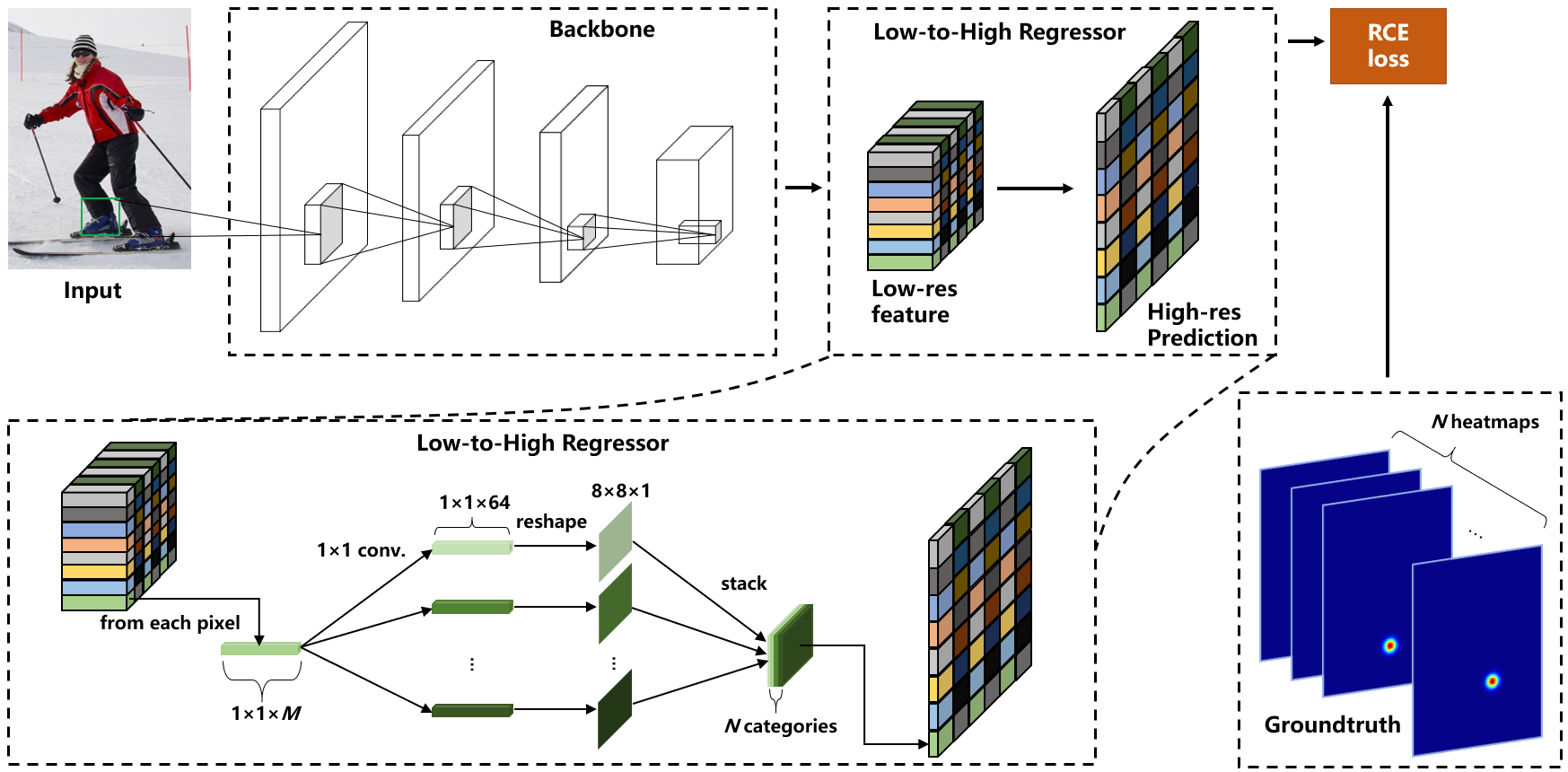}
    \caption{The overview of FasterPose. The input image is passed through a backbone, and then through the LHR module, and finally approaches the heatmap regression step that supervised by the RCE Loss. 
    Each group of 1$\times$1 convolution contains $L^2$ kernels, and $L$ is the upsampling ratio (\textit{e.g.}~ $L=8$ in this figure).
    The number of groups equals to the categories of keypoint.
    }
    \label{fig:pipeline}
\end{figure*}

\subsection{Fast Paradigm}
\label{sec:method:fast}

The objective of heatmap-based human pose estimation is to accomplish the regression of output heatmaps that correspond to each keypoint.
Typically, the input image is passed through a backbone that consists of high-to-low resolution subnetworks, and then through the subsequent resolution-raising subnetworks, and finally approaches the heatmap regression stage.
The high-to-low stage aims to generate the LR and highly-semantic representations, and the low-to-high stage subsequently increases the resolution to support the accurate keypoint localization in final~\cite{bulat2016human,chen2018cascaded,hu2016bottom,xiao2018simple,newell2016stacked,tang2018deeply}.
Therefore, this set of methods highly depends on the learning and inference of HR representation, suffering from the problem of heavy computational cost that in proportion to the resolution.
An effective alternative consists in cutting out the low-to-high stage, and inferring the heatmap from the LR representation directly. Without any modification of the remaining steps, the network is built in a straightforward way, while the critical challenge comes from the regressor which recovers the spatial resolution from the LR representation to the output heatmap and simultaneously regresses the prediction on the heatmap. 
Previously, SimpleBaseline~\cite{xiao2018simple} deals with this challenge by using multiple layers of deconvolution, which improves the regression accuracy but also leads to significant augmentation of computational cost. For example, when the backbone (\textit{i.e.},~ the high-to-low stage) is ResNet-50, the additional FLOPs that brought by the deconvolution layers will take 59\% budget of the entire network.

Rather than sticking to the framework of upsampling from spatial-neighboring site, we assume that the LR features have great capacity of semantic information along the channel dimension, and we show how to make use of the cross-channel abundant semantic information for regressing the heatmap via 1$\times$1 convolutions. 
In other words, the high-level semantic features are sufficient to locally infers the prediction in the low-frequency domain.
This idea encourages us to build the scheme of LHR (\textbf{Low-to-high Regressor}) in FasterPose.
Specifically, LHR applies a number of groups of 1$\times$1 convolution following the backbone; each group contains the kernels of which the number equals to the desired square upsampling ratio.
Without loss of generality, we take an example that the ratio is $L$, and the target has $N$ keypoints, then LHR employs $N$ groups, each of which owns $L^2$ kernels of $1\times1\times M$, where $M$ denotes the channel number of the LR feature maps. So, each group is in charge of projecting the LR feature to the heatmap with $L^2$ channels, or equally the heatmap with $L$ upsampling ratio in width and height; the total $N$ groups enable to output $N$ heatmaps corresponding to the $N$ keypoints, respectively.

\begin{figure*}[t]
    \centering
    \includegraphics[width=0.8\linewidth]{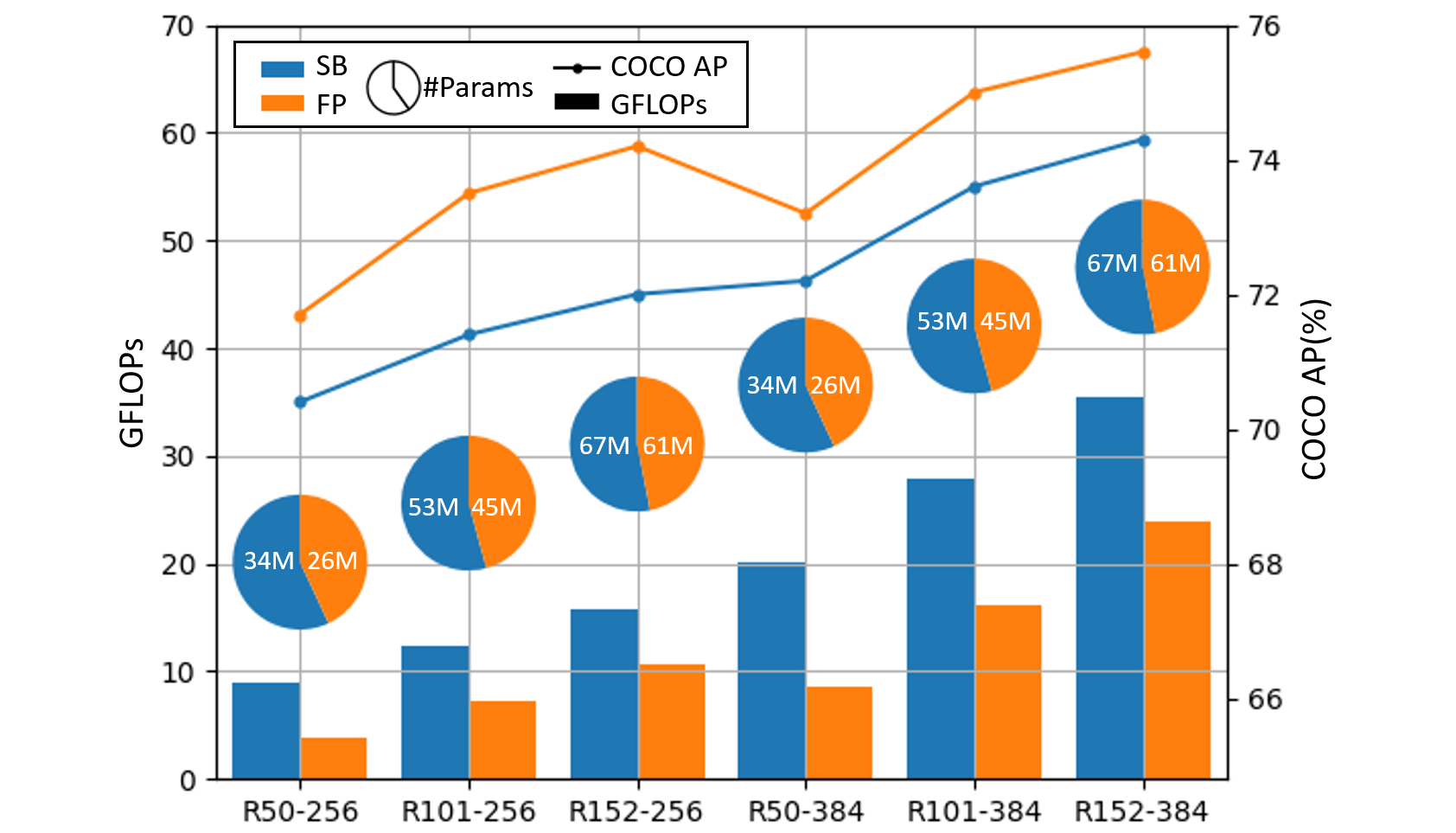}
    \caption{The detailed comparison between FasterPose (FP) and SimpleBaseline (SB) with various backbone architectures and input size, in terms of FLOPs, model size and COCO AP.
    The horizontal axis indicates the choice of backbone and input size. For example, ``R50-256'' denotes the ResNet-50 with input size of 256$\times$192 RGB.
    }
    \label{fig:AP_GFLOPs}
\end{figure*}

LHR is currently the simplest structure to infer HR heatmaps from LR features for pose estimation.
Its parameter number only takes $M \times N \times L^2$, while the deconvolutions in SimpleBaseline carry parameters of $(M+2\times F) \times F \times K^2 + F \times N$, where $K$ and $F$ are the kernel size and kernel number of the deconvolutions.
Resorting to LHR, we can reduce 79\% of the regressor weights and 25\% of the total weights compared with the counterpart subject to ResNet-50, $K=4$ and $F=256$ in SimpleBaseline.
Fig.~\ref{fig:AP_GFLOPs} shows that LHR significantly reduces the FLOPs as well.

It is noteworthy that the PixelShuffle method proposed by ESPCN~\cite{shi2016real},
which is initially designed for the super-resolution task, is similar to LHR in terms of the LR inputs. ESPCN adopts an 3$\times$3 convolution to generate the features with $ (L^2, W, H)$ shape firstly, and then shuffle to the up-sampled maps of $ (1, W\times L, H\times L)$. Such routine cannot be used in most cases of pose estimation without customization, since the number of output channel of backbone (generally 512 or 2048) divided by the category of human keypoint (\textit{e.g.}~ 17 in COCO setting) does not match the multiples of the number $L^2$; instead, LHR gets rid of such limit, and can be implemented with any off-the-shelf architectures for pose estimation. Besides, by taking the advantage of abundant local information in the LR features, LHR accomplishes the mapping from LR feature to heatmap with 1$\times$1 convolution of which the computational cost is only 1/9 of PixelShuffle.

\subsection{RCE Loss}
\label{sec:method:rce}

\subsubsection{The drawback of MSE for pose estimation}
\label{sec:method:rce:mse}

The MSE loss is employed by most of the heatmap regression methods to supervise the training of the backbone and regressor. 
It minimizes the pixel-wise discrepancy between the predicted heatmap and the ground truth heatmap.
While we adopt MSE for training FasterPose, we find the network suffers from slow convergence and inferior accuracy (Fig.~\ref{fig:training_curve}) compared with the HR-style counterpart.
To identify the reason, we inspect the distribution of the heatmap values (Table~\ref{tbl:distribution_of_samples}). 
It is obvious that, for each pose to be estimated, the true keypoints, whose values are greater than zero in the ground truth heatmap, are distributed sparsely in the spatial dimension, while the remaining massive background pixels are with zero value. 
The parameters of the heatmap regressor are significantly reduced when the feature becomes LR, and most of the parameters are dedicated to the regression towards zero value in the background under the supervision of MSE.
This is the reason that causes the convergence issue with respect to the positive samples (\textit{a.k.a.}~ keypoints) which is precisely the primary objective of pose estimation. 
Moreover, the inferior accuracy is also caused by the reason that MSE is inconsistent with the evaluation metric of pose estimation. For example, a minor mis-prediction in the keypoint site affects the test accuracy (\textit{e.g.}~ AP) remarkably, while a minor mis-prediction in background barely matters; on the other hand, both the predictions in background and keypoint site contribute equally to the MSE value.

\begin{figure}[t]		
\centering 
\label{fig:loss}
\subfigure[]
{\includegraphics[width=0.35\linewidth]
{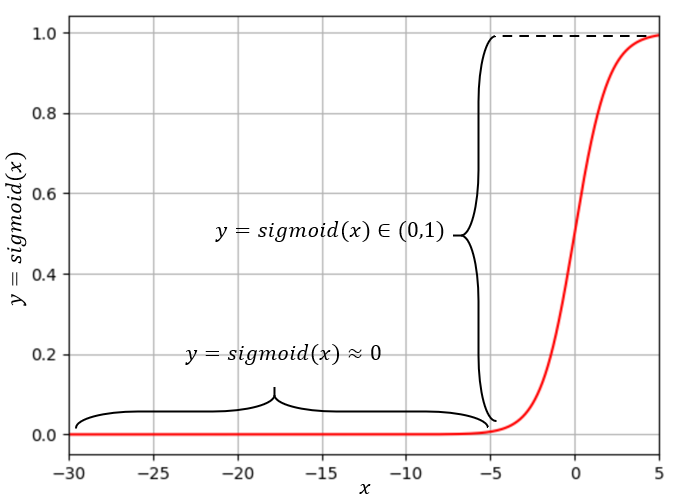}
\label{fig:loss_a}}
\subfigure[]
{\includegraphics[width=0.332\linewidth]
{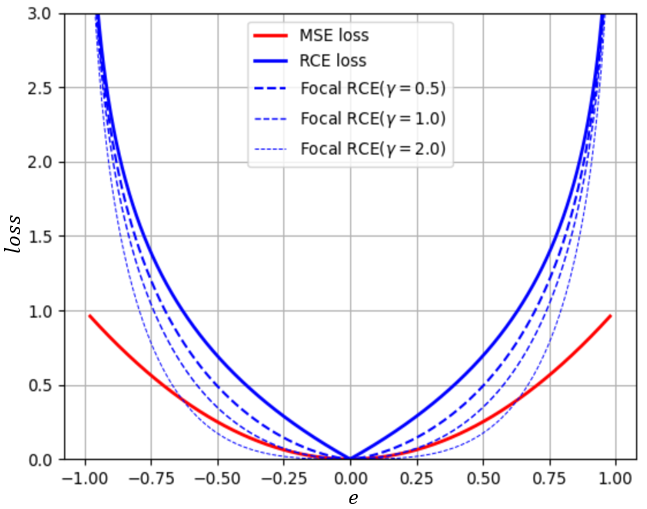}
\label{fig:loss_b}}
\subfigure[]
{\includegraphics[width=0.35\linewidth]
{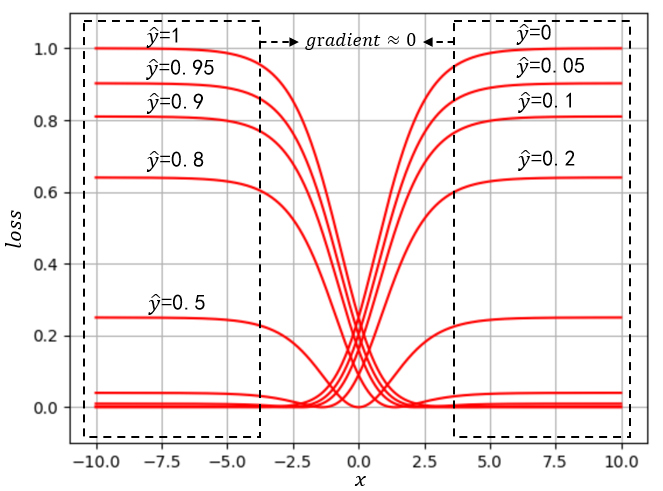}
\label{fig:loss_c}}
\subfigure[]
{\includegraphics[width=0.34\linewidth]
{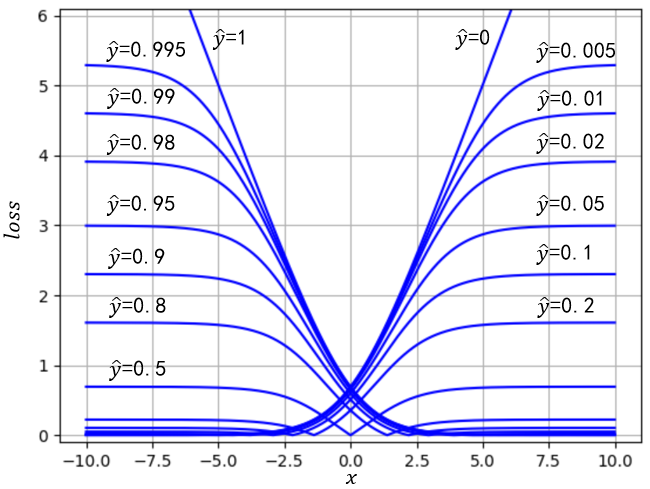}
\label{fig:loss_d}}
\caption{
Analysis of loss function. (a) The sigmoid function, where $x$ (horizontal axis) represents the regression value, and $ y = sigmoid(x) $ squeezes most values to 0 and 1.
(b) MSE and RCE with respect to the error $ e = y - \hat y $, where $\hat y$ denotes the ground truth.
(c) MSE with respect to $x$, whose gradient vanishes in most range of $x$.
(d) RCE alleviates the gradient vanishing problem.
Best viewed in color.
}
\end{figure}

To solve this problem, the sigmoid operation is a more suitable choice than the Euclidean metric. 
As shown in Fig.~\ref{fig:loss_a}, the regression value $x$ is squeezed to 0 or 1 by the sigmoid mapping. For example, the values $x \in \left( -\infty, -5 \right]$ are mapped sufficiently close to zero.
Thus, the regressor does not have to fit the background with large portion of parameters.
Instead, the regressor is released for the primary objective, \textit{i.e.},~ the regression towards positive samples.
A side affect, however, is raised when MSE applied to $y = sigmoid(x)$: the gradient rapidly vanishes in most range of $x$ (Fig.~\ref{fig:loss_c}). We tackle this problem with the RCE loss.

\begin{table}[t]
	\setlength{\tabcolsep}{0.1cm}
	\begin{center}
		\caption{
		    Statistics of the positive (P)  and negative (N) samples in heatmap.
		    Most of the pixels in heatmap are negative samples.
			$t$ is the variance of Gaussian diffusion.
		}
		\label{tbl:distribution_of_samples}
		\begin{tabular}{c | c | c |c  |c }
			\hline
			Heatmap size & $t$ & P & N  & Ratio of P:N \\
			\hline \hline
			64$ \times $48 & 2.0 & 113 & 2,959 & 1:26.2
			\\ 
			96$ \times $72 & 3.0 & 261 & 6,651 & 1:25.5
			\\ 
			\hline
		\end{tabular}
	\end{center}
\end{table}

\subsubsection{Formulation of RCE}
\label{sec:method:rce:formulation}

There are only two prior works~\cite{Moon2019PoseFix,papandreou2017towards} that use the CE loss for training human pose estimation.
The major challenge is that the ground truth heatmap consists of real numbers that ranges from 0 to 1, which is conflict to the binary objective of CE loss:
\begin{equation}
\label{eq:CE loss}
CE (y, \hat y) = 
\begin{cases}
- log (y) &  \text{if $\hat y = 1,$} \\ 
- log (1-y) &  \text{if $\hat y = 0, $} 
\end{cases}
\end{equation}
where $\hat y \in \{ 0, 1\}$ is the objective, and $y \in \left[ 0,1 \right]$ is the prediction. 
G-RMI~\cite{papandreou2017towards} and PoseFix~\cite{Moon2019PoseFix} dispose a roundabout that quantifies all the positive samples to 1. 
However, such quantifing practice will introduce bias away from the ground truth and bring sub-optimal training (Table~\ref{tbl:loss}).
To deal with this obstacle, we reshape the CE loss to a novel formulation, RCE, which is able to supervise the training toward real numbers between 0 and 1.
Specifically, given the prediction error $e = y - \hat y$, the RCE loss is formulated by 
\begin{equation}
\label{eq:modified CE loss}
RCE (y, \hat y) = RCE (e) = - log (1-|e|).
\end{equation}
The RCE loss has three advantages. 
Firstly, the objective can be generalized to the real numbers $\hat y \in \left[ 0, 1\right]$, thereby the pose estimation training can benefit from RCE loss. One may notice that the CE loss becomes a special case of RCE if the objective $\hat y$ is restricted to the integers of 0 and 1. 
Secondly, the RCE loss alleviates the gradient vanishing issue that caused by the sigmoid mapping (Fig.~\ref{fig:loss_d}).
Thirdly, the loss value goes to infinity when $|e|$ approaches 1 (Fig.~\ref{fig:loss_b}), and such trend provides the prerequisite for learning from the hard samples with large loss values. In contrast, the MSE loss cannot yield such large loss value on hard samples within the finite support of $e$.
In the following part, we will take this advantage and further develop the RCE loss with the hard sample learning.

\subsubsection{RCE against hard sample}
\label{sec:method:rce:focal}
Hard sample (\textit{a.k.a.}~ outlier pixel) is a common issue in training pose estimation network.
The common practice for hard sample learning is to introduce a weighting factor in the loss formulation, where the weighting factor is set proportional to the hard extent.
Following the common practice, we adopt the absolute prediction error $|e|$ as the measurement of hard extent. 
Owing to the advantage described above, when $|e|$ increases, the large loss value of RCE provides the support of hard sample emphasis.
The final RCE loss is developed in a focal style:
\begin{equation}
\label{eq:focal_rce_loss}
RCE (e) = - |e|^\gamma log (1-|e|),
\end{equation}
where the power $\gamma$ is a hyperparameter for modulating the hard sample weighting. In the experiments, we find that a trivial setting of $\gamma =1.0$ is enough to bring consistent improvement.
Also, as suggested by Focal loss~\cite{Lin2017Focal}, a further boost can be obtained by modulating the weighting factor with respect to positive and negative, \textit{e.g.}~ adopting $\alpha|e|^\gamma$ for positive samples, and $(1-\alpha) |e|^\gamma$ for negative samples.
It should be noted that the RCE loss is not limited with any specific focal routine, but is friendly to various formulation for hard sample learning.

\section{Experiments}
\label{sec:exp}

\subsection{Experimental Setting}
\label{sec:exp:setting}

\subsubsection{Datasets}
\label{experiments:datasets}
We utilized two popular human pose estimation datasets, COCO~\cite{lin2014microsoft} and MPII~\cite{andriluka20142d}. The COCO keypoint dataset~\cite{lin2014microsoft} presents naturally challenging imagery data with various human poses, unconstrained environments, different body scales and occlusion patterns. The total objective involves both detecting the person instances and localizing the keypoints, while this work focus on the latter one.
It contains 200k images and 250k person samples. Each person instance is labeled with 17 keypoints. The annotations of training and validation sets are publicly benchmarked. In the evaluation, we follow the commonly used protocol of train2017/val2017/test-dev2017 split. The MPII human pose dataset~\cite{andriluka20142d} contains 40k person samples, each of which is labeled with 16 keypoints. We follow the standard train/val/test split as in~\cite{tompson2014joint}. 

\subsubsection{Training details}
\label{experiments:training}
The weights of the backbone part are initialized with the publicly released ResNet model that pre-trained on the ImageNet dataset~\cite{russakovsky2015imagenet}. The weights are updated by the Adam optimizer with the mini-batch size of 128. The initial learning rate is set to 1$\times 10^{-3}$, and divided by 10 at 90 and 120th epoch, respectively. The training process is terminated within 140 epochs.
We perform data augmentation including scaling ($\pm$ 35\%), rotation ($\pm45^{\circ}$), and flip on COCO. Following~\cite{wang2018mscoco}, half body data augmentation is involved.
We use two different input sizes (256$\times$192, 384$\times$288) in our experiments, and adopt the same data preprocessing as in~\cite{xiao2018simple}.
Samples from the MPII are augmented during training by using the same scheme as in Hourglass~\cite{newell2016stacked}.
In addition, we transform the plain regression output $x$ with $sigmoid(x)$ when using RCE loss. In this case, we fix $\alpha$=0.7, $\gamma$=1.0 in Eq.~\ref{eq:focal_rce_loss} according to the validation result, and find it stably performs well throughout the experiments.

\subsubsection{Evaluation}
\label{experiments:evaluation}
We use the standard average precision (AP) on COCO and the percentage of correct keypoints (PCK) on MPII to evaluate the accuracy.
The model efficiency is assessed in multiple terms, including the FLOPs, the parameter number.
and the inference time per image measured on a single NVIDIA 2080ti GPU.
The top-down paradigm is used on COCO: it detects the person instance with a human detector, and then predicts the keypoints. We use the same person detectors~\cite{faster2015towards} provided by SimpleBaseline for both validation set and test-dev set.
Also the same with SimpleBaseline, we compute the heatmap by averaging the heatmap of the original and  flipped images.
Then we implement a quarter offset from the highest response to the second highest one to get the locations of keypoints. The final confidence score is obtained by the multiplication of the averaged score of keypoints and the bounding box score. 
The testing procedure of MPII is generally the same to that in COCO except that we follow the standard testing strategy with the provided person boxes instead of detected person boxes. 

\subsubsection{Contrast methods}
The success of SimpleBaseline~\cite{xiao2018simple} has provided the prior knowledge on how to design an efficient network for human pose estimation.
Inspired by this design, LPN~\cite{zhang2019simple} moves further towards the lightweight models. 
Apart from them, the remaining pose estimation methods in literature pursue the accuracy with heavy weights, and seldom notices the efficiency issue. 
Therefore, our experiments focus on the comparison with SimpleBaseline and LPN, in order to validate the advantages of FasterPose with respect to the efficiency.
Besides, many state-of-the-art methods are involved in the comparison experiments, such as HRNet~\cite{sun2019deep}, RSN~\cite{sun2019deep}, G-RMI~\cite{papandreou2017towards}, Integral Pose Regression~\cite{sun2018integral}, RMPE~\cite{fang2017rmpe}, CPN~\cite{chen2018cascaded}, Lite-HRNet~\cite{Yulitehrnet21}, \textit{etc}.

\subsection{COCO Keypoint Detection}
\label{sec:exp:coco}

\subsubsection{Results on validation set}
\label{experiments:coco_val}
The comparison between SimpleBaseline and FasterPose is shown in Table~\ref{tbl:sb_val}. It is obvious that FasterPose reduces the computational cost, model size and inference time dramatically, while improving the accuracy. 
FasterPose benefits from: 
(1) the LHR module directly uses LR features for regression, greatly reduces the computational cost; 
(2) the RCE loss significantly improve the convergence.
Given the input size of either 256 $\times$192 or 384$ \times $288, FasterPose consistently yields higher AP than the counterpart with various backbone architectures.
Meanwhile, the FLOPs, parameter number, and inference time are significantly reduced by FasterPose.
In order to verify the generalization ability of FasterPose, we also carried out experiments on LPN~\cite{zhang2019simple}. 
Different with SimpleBaseline, LPN~\cite{zhang2019simple} replace the standard bottleneck blocks used in the backbone with a lightweight bottleneck blocks when downsampling. During the upsampling process, LPN replace each deconvolutional layer with a combination of a group deconvolutional layer. Hence, LPN has much less model size and computational complexity than other top-performing networks.
The results of the comparison with LPN and FasterPose are shown in the Table~\ref{tbl:sb_val}, FasterPose achieves the AP increase of 0.8\% (67.6$\% \to$ 68.2\%), and only needs 80\% FLOPs than that of LPN. 

\begin{table}[t]
	\setlength{\tabcolsep}{0.1cm}
	\begin{center}
		\caption{
		    Comparison of LPN, SimpleBaseline and FasterPose on COCO validation set. 
		    The results show that FasterPose has consistent advantage of accuracy and efficiency with various backbone architectures and input size.
		}
		\label{tbl:sb_val}
		\begin{tabular}{l | c | l |c | c | c | c}
			\hline
			Method & Input Size &  Backbone &  GFLOPs & \#Params & Infer. time(ms) & AP(\%) \\
			\hline \hline
			LPN & 256 $\times$ 192 & lpn-50 & 1.0 & 2.91M & 4.8 & 67.6 \\
			FasterPose & 256 $\times$ 192 & lpn-50 & \bf0.8 & \bf2.90M & \bf4.6 & \bf68.2 \\
			\hline
			\hline
			SimpleBaseline & 256 $\times$ 192 & ResNet-50 & 9.0 & 34.0M & 7.5 & 70.4 \\
			SimpleBaseline & 256 $\times$ 192 & ResNet-101 & 12.4 & 53.0M & 15.0 & 71.4 \\
			SimpleBaseline & 256 $\times$ 192 & ResNet-152 & 15.8 & 68.6M & 20.9 & 72.0 \\
			FasterPose & 256 $\times$ 192 & ResNet-50 & \bf3.8 & \bf25.7M & \bf6.4 & \bf71.7 \\
			FasterPose & 256 $\times$ 192 & ResNet-101 & \bf7.2 & \bf44.7M & \bf13.0 & \bf73.5 \\
			FasterPose & 256 $\times$ 192 & ResNet-152 & \bf10.6 & \bf60.4M & \bf18.7 & \bf74.2 \\
			\hline
			\hline
			SimpleBaseline & 384 $\times$ 288 & ResNet-50 & 20.2 & 34.0M & 9.4 & 72.2 \\
			SimpleBaseline & 384 $\times$ 288 & ResNet-101 & 27.9 & 53.0M & 17.5 & 73.6 \\
			SimpleBaseline & 384 $\times$ 288 & ResNet-152 & 35.5 & 68.6M & 24.1 & 74.3 \\
			FasterPose & 384 $\times$ 288 & ResNet-50 & \bf8.6 & \bf25.7M & \bf7.5 & \bf72.7 \\
			FasterPose & 384 $\times$ 288 & ResNet-101 & \bf16.2 & \bf44.7M & \bf14.3 & \bf75.0 \\
			FasterPose & 384 $\times$ 288 & ResNet-152 & \bf23.9 & \bf60.4M & \bf21.5 & \bf75.6 \\
			\hline
		\end{tabular}
	\end{center}
\end{table}

\begin{table*}[t]
\setlength{\tabcolsep}{0.1cm}
\begin{center}
\caption{
Comparison with state of the art on COCO test-dev set. FasterPose reduces the computational cost dramatically, while keeping the accuracy competitive. 
Note that RSN involves the heavy complexity that far outweighs the others. The accuracy is reported in percentage.
}
\label{tbl:STOA_COCO}
\resizebox{\columnwidth}{!}{
\begin{tabular}{l|l|c|c|c|c|c|c|c|c|c|c}
\hline
Method & Backbone & lightning & Input size & \#Params & GFLOPs & AP& AP$^{50}$ & AP$^{75}$ & AP$^{M}$ & AP$^{L}$ & AR  \\
\hline \hline
CPN~\cite{chen2018cascaded} & ResNet-Inc & - & 384$\times$288 & - & -
& 72.6 & 86.1 & 69.7 & 78.3 & 64.1 & -
\\
CPN~\cite{chen2018cascaded} (ensemble) & ResNet-Inc & - & 384$\times$288 & - & -
& 73.0 & 91.7 & 80.9 & 69.5 & 78.1 & 79.0 
\\
RSN~\cite{sun2019deep} & 4$\times$ RSN-50  & - & 384$\times$288 & 111.8M & 65.9 & 78.6 & 94.3 & 86.6 & 75.5 & 83.3 & 83.8
\\
G-RMI~\cite{papandreou2017towards}
& ResNet-101 & - & 353$\times$257 & 42.6M & 57.0 
& 64.9 & 85.5 & 71.3 & 62.3 & 70.0 & 69.7
\\	G-RMI~\cite{papandreou2017towards}+extra data &  ResNet-101 & - & 353$\times$257 & 42.6M & 57.0 
& 68.5 & 87.1 & 75.5 & 65.8 & 73.3 & 73.3
\\	SimpleBaseline~\cite{xiao2018simple} & ResNet-152 & - & 384$\times$288 & 68.6M & 35.6
& 73.7 & 91.9 & 81.1 & 70.3 & 80.0 & 79.0
\\
HRNet~\cite{sun2019deep} & HRNet-W48 & - & 384$\times$288 & 63.6M & 32.9
& 75.5 & 92.5 & 83.3 & 71.9 & 81.5 & 80.5
\\
RSN~\cite{sun2019deep} & 4$\times$ RSN-50 & - & 256$\times$192 & 111.8M  & 29.3 & 78.0 & 94.2 & 86.5 & 75.3 & 82.2 & 83.4
\\
RMPE~\cite{fang2017rmpe}
& PyraNet & - & 320$\times$256 & 28.1M & 26.7
& 72.3 & 89.2 & 79.1 & 68.0 & 78.6 & -
\\
HRNet~\cite{sun2019deep} & HRNet-W32 & - & 384$\times$288  & 28.5M & 16.0
& 74.9 & 92.5 & 82.8 & 71.3 & 80.9 & 80.1
\\
Integral Pose Regression~\cite{sun2018integral}
& ResNet-101 & - & 256$\times$256 & 45.0M & 11.0
& 67.8 & 88.2 & 74.8 & 63.9 & 74.0 & -
\\	SimpleBaseline~\cite{xiao2018simple}
& ResNet-50 & - & 256$\times$192 & 34.0M & 9.0
& 69.7 & 90.9 & 78.5 & 67.4 & 75.0 & 75.9
\\
Lite-HRNet~\cite{Yulitehrnet21} & Lite-HRNet-18 & \checkmark & 256$\times$192 & 1.13M  & 0.21 & 64.1 & 88.7 & 72.0 & 61.6 & 69.0 & 70.2
\\
\hline
\hline
FasterPose & ResNet-50 & - & 256$\times$192 &  25.7M &  3.8
& 70.8 & 91.3 & 78.8 & 67.2 & 76.8 & 76.4
\\
FasterPose & Lite-HRNet-18 & \checkmark & 256$\times$192 & 1.13M  & 0.21 & 64.5 & 89.0 & 72.4 & 62.2 & 69.3 & 70.6
\\ \hline
\end{tabular}
}
\end{center}
\end{table*}

\subsubsection{Results on test-dev set}
\label{experiments:coco_test}
We further evaluate the proposed method on the COCO test-dev set, and make the comparison with the state-of-the-art methods. Table~\ref{tbl:STOA_COCO} compares the accuracy results of state-of-the-art methods and FasterPose.
Following the conventional routine, we use the human bounding boxes provided by~\cite{sun2019deep}.
We can find that SimpleBaseline is the most efficient among the prior networks that do not use any weight-lightning strategy (\textit{i.e.},~ weight-lightning blocks like MobileNet~\cite{sandler2018mobilenetv2} or ShuffleNet~\cite{zhang2018shufflenet}), while FasterPose is able to further improve the efficiency with competitive accuracy.
FasterPose achieves the AP increase of 1.1\% (69.7\% $\to$ 70.8\%) compared to SimpleBaseline, and only needs 42.2\% FLOPs than that of SimpleBaseline. 
Beyond the SimpleBaseline, weight-lightning practice~\cite{zhang2018shufflenet} currently helps HRNet greatly reduce model complexity, \textit{a.k.a.}~ Lite-HRNet~\cite{Yulitehrnet21}.
So, we adopt Lite-HRNet as backbone to establish an extra version of FasterPose, and yield higher accuracy than the Lite-HRNet (bottom line in Table~\ref{tbl:STOA_COCO}). As no need of LHR module in this case, it indicates the advantage of RCE loss.

\subsection{MPII Keypoint Detection}
\label{sec:exp:mpii}

Table~\ref{tbl:MPII} shows the accuracy in terms of PCKh@0.5, and the FLOPs of SimpleBaseline and FasterPose.
Overall, we obtain a similar comparison result with that on COCO. In particular,  
FasterPose exhibits the superior accuracy over the counterparts in every backbone configuration, while constantly reducing the FLOPs. 
It is noteworthy that MPII provides a much smaller training dataset than COCO, suggesting that our method generalizes well across training data sizes.

\begin{table}[t]
	\setlength{\tabcolsep}{0.1cm}
	\begin{center}
	\caption{
	    Comparison of FasterPose and SimpleBaseline (PCKh@{0.5} with single-scale test) on the MPII validation set. The input size is 256 $\times$ 256 RGB.
	}
	\label{tbl:MPII}
			\begin{tabular}{l | l | c | c | c | c | c | c | c | c | c}
				\hline
				Method & Backbone & GFLOPs & Hea & Sho & Elb & Wri & Hip & Kne & Ank & Mean \\
				\hline \hline
                SimpleBaseline & ResNet-50 & 12.0 & 96.4 & 95.3 &  89.0 &  83.2 & 88.4 & 84.0 &  79.6 & 88.5 \\
                SimpleBaseline & ResNet-101 & 16.5 & 96.9 & 95.9 & 89.5 & 84.4 & 88.4 & 84.5 & 80.7 & 89.1 \\
                SimpleBaseline & ResNet-152 & 21.0 & 97.0 & 95.9 & 90.0 & 85.0 & 89.2 & 85.3 & 81.3 & 89.6 \\
                \hline
                \hline
                FasterPose & ResNet-50 & \bf 5.1 & 96.5 & 95.4 & 88.9 & 83.5 & 88.4 & 84.1 & 79.7 & \bf88.6 \\
                FasterPose & ResNet-101 & \bf 9.6 & 96.7 & 95.9 & 89.5 & 84.4 & 88.6& 85.8 & 81.0 & \bf 89.3 \\
                FasterPose & ResNet-152 & \bf 14.1 & 96.8 & 95.7 & 90.1 & 85.0 & 89.2 & 86.1 & 82.7 & \bf 89.8 \\
                \hline

			\end{tabular}
	\end{center}
	
\end{table}

\subsection{Ablation Study}
\label{sec:exp:ablation}

\subsubsection{Influence of feature resolution}
We study how the representation resolution affects the accuracy of human pose estimation.
First, we verify the performence of SimpleBaseline with the various feature resolutions. The input image of 256$\times$192 RGB is passed through the ResNet-50 backbone to extract the LR features of 8$\times$6.
Then, we employ one, two and three layers of deconvolution to generate features with resolutions of 16$\times$12, 32$\times$24 and 64$\times$48, respectively. As a result, we obtain four representations with different resolutions.
Subsequently, they are used to regress the output heatmap.
The comparison is summarized in Table~\ref{tbl:resolution}, including the evaluation of AP on COCO, the FLOPs, and the models size.
It is observed that when the resolution of the feature increases, the computational cost of the model also increases greatly. The improvement of accuracy, however, is only 0.1\%. 
This observation confirms that the high efficiency of LR features is valid for human pose estimation, although it has been neglected in the past.

\begin{table}[t]
	\setlength{\tabcolsep}{0.1cm}
	\begin{center}
	    \caption{
	        The ablation study on the feature resolution.
			With the increase of resolution, the accuracy barely changes, while the computational cost significantly increases.
			The backbone is ResNet-50, and the input size is 256 $\times$ 192 RGB. The same data augmentation as SimpleBaseline are used.
		}
		\label{tbl:resolution}
			\begin{tabular}{c | c | c | c | c}
				\hline
				Feature size & AP (\%) & AP$^{50}$ (\%) & GFLOPs & \#Params\\
    			\hline \hline
    			8$\times$6
    			& 70.3 & 88.6 & \bf 3.8 & \bf 25.6M
    			\\ 
    			16$\times$12
    			&  70.4 & 88.7 & 5.2 & 31.9M
    			\\ 
    			32$\times$24
    			& 70.4 & 88.7 & 6.0 & 33.0M
    			\\ 
    			64$\times$48
    			& 70.4 &  88.6 & 9.0 & 34.0M
    			\\ 
    			\hline
			\end{tabular}
    \end{center}
\end{table}

\begin{figure}[t]
    \centering
    \includegraphics[scale=0.3]{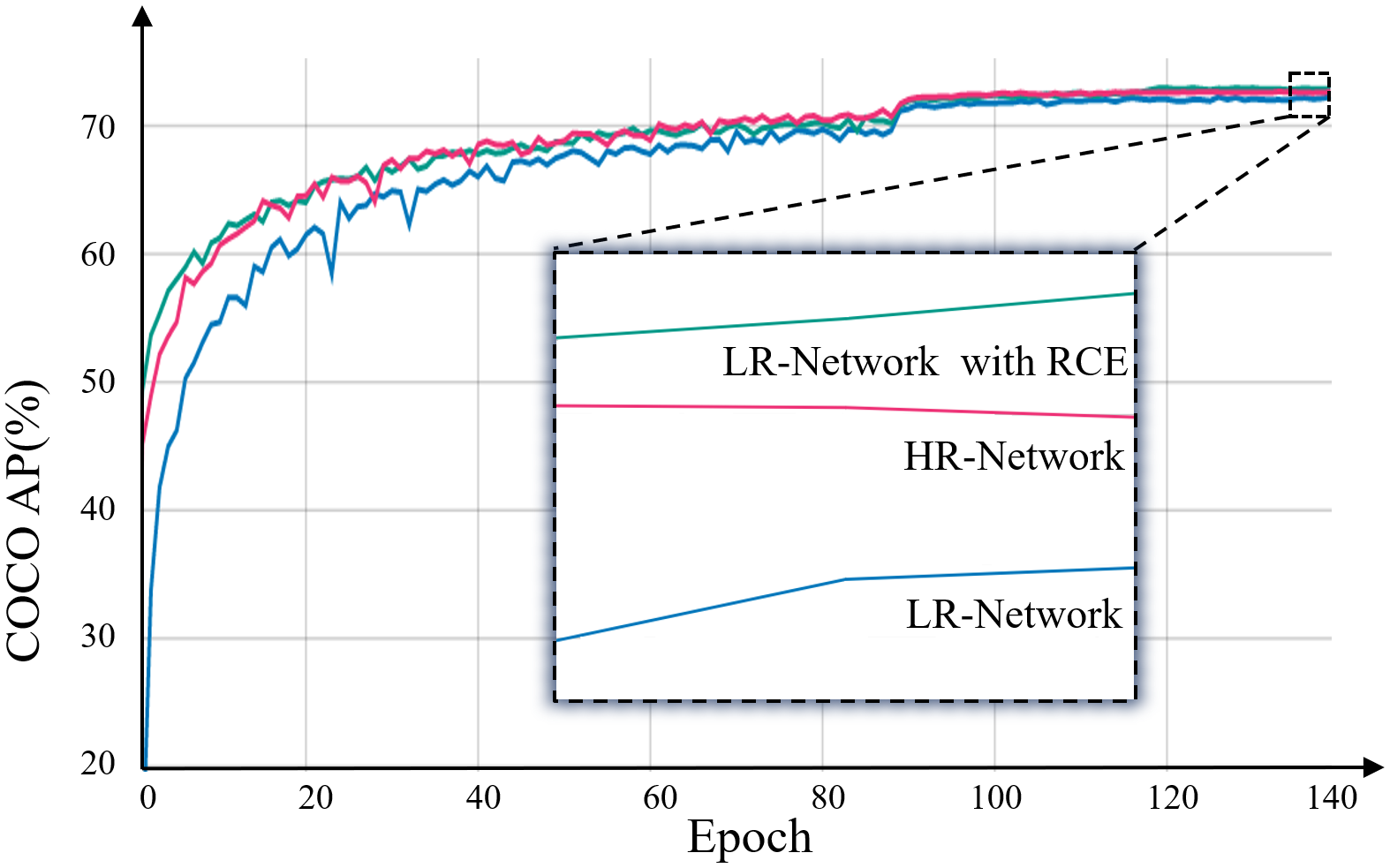}
    \caption{
    The convergence of the HR-based and LR-based networks, measured by the AP on COCO validation.
    The green curve denotes the training with RCE supervision, while the others are supervised by MSE.
    All of them have the same backbone (ResNet-50) and input resolution (256$\times$ 192 RGB).
    The HR-based network employs three deconvolution layers to recover the HR features, while the LR-based network employs the LHR module to regress the heatmap from the LR features directly.
    It is obvious that the convergence of LR-based network (blue curve) is slower than that of the HR-based network (red curve). 
    The RCE loss promotes the LR-based network in terms of convergence and accuracy (green curve).
    }
    \label{fig:training_curve}
\end{figure}

\subsubsection{Advantage of RCE loss}
\label{experiments:rce_loss}
Based on the efficient architecture of FasterrPose, the next goal is to further promote the spatial accuracy for pose estimation. We visualize the convergence process and find that the convergence of LR-based network is slower than that of HR-based network, therefore resulting in inferior performance. This verifies the conclusion: the reduction of parameters over the feature layers will cause the model unable to accurately learn from the samples with imbalance classes.
To deal with this problem, the RCE supervision is developed to promote the training convergence and the final accuracy.

As mentioned in Section~\ref{sec:method:rce}, the standard CE loss can only supervise the learning with ground truth of 0-1 binary values, while RCE loss can be applied to the heatmap regression learning with the ground truth of continuous real values. In this experiment, we compare the RCE loss with not only the CE loss, but also the MSE loss.
The APs are shown in Table~\ref{tbl:loss}. It can be clearly observed that the RCE loss outperforms both MSE loss and CE loss.
Besides, in Fig.~\ref{fig:training_curve}, the green curve shows the advantage of RCE loss for improving the convergence, as well as the accuracy.

\begin{table}[t]
	\setlength{\tabcolsep}{0.1cm}
	\begin{center}
	    \caption{
			The RCE supervision consistently improves the accuracy compared with MSE and CE on COCO validation set. The model is FasterPose (ResNet-50), the input size is 256$\times$192 RGB, The accuracy is reported in percentage.
		}
		\label{tbl:loss}
			\begin{tabular}{l | c | c | c | c | c | c}
				\hline
				Loss  & AP  & AP$^{50}$ & AP$^{75}$  & AP$^{M}$  & AP$^{L}$  & AR  \\
				\hline \hline
    			MSE & 70.3 & 88.6 & 78.8 & 67.3 & 77.8 & 76.6
    			\\ 
    			CE (one-hot)& 69.7 & 89.1 & 78.1 & 67.3 & 77.2 & 76.2
    			\\ 
    			CE (mask) 
    			& 70.0 & 88.9 & 76.9 & 66.6 & 76.5 & 76.6
    			\\ 
    			RCE & \bf 70.9 & \bf 89.5 & \bf 78.9 & \bf 67.8 & \bf 77.9 & \bf 76.8
    			\\ 
    			\hline
			\end{tabular}
    \end{center}
\end{table}

\section{Conclusion}
\label{sec:conclusion}

In this work, we present a novel design paradigm, named FasterPose, to accomplish the efficient network for human pose estimation, and a novel loss function for effectively training such efficient network.
To the best of our knowledge, it is the first attempt to analyze the impact of feature resolution for human pose estimation. 
Based on the analysis, FasterPose is devised with the design of LR feature and thereby largely reduces the computational cost.
Furthermore, we study the convergence behavior of LR based networks, and formulate a novel loss function for accelerating its convergence and raising its accuracy.
Our method facilitate the low-latency and low-energy-budget application as required in the non-GPU scenarios. 
The extensive experiments verify the effectiveness and efficiency of FasterPose.

{\small
	\bibliographystyle{ACM-Reference-Format}
	\bibliography{sample-base}
}

\end{document}